\documentclass[runningheads]{llncs}
\usepackage[T1]{fontenc}

\usepackage{comment}
\usepackage{blindtext}
\usepackage{dirtytalk}
\usepackage{array}
\usepackage{authblk}
\usepackage{url}
\usepackage[noadjust]{cite}
\usepackage{graphics}
\usepackage{graphicx}
\usepackage{epsfig}
\usepackage{amssymb}
\usepackage{amsmath}
\usepackage{breqn}
\usepackage[ruled,vlined]{algorithm2e}

\usepackage{mathtools}
\usepackage{lscape}
\usepackage{pdflscape}
\usepackage{longtable}
\usepackage{makecell}
\usepackage{lettrine}
\usepackage{caption}
\usepackage{supertabular,booktabs}

\usepackage[]{hyperref}
\usepackage[table,xcdraw]{xcolor}
\usepackage{tabularx}
\usepackage{multirow}
\usepackage{subcaption}
\usepackage{orcidlink}

\usepackage{color}

\begin{document}
\title{Enhancing Performance for Highly Imbalanced Medical Data via Data Regularization in a Federated Learning Setting}

\titlerunning{Federated Data Regularization for Highly Imbalanced Medical Data}

\author{Georgios Tsoumplekas\inst{1}\footnote{Corresponding author.}\orcidlink{0009-0004-4943-3381} \and
Ilias Siniosoglou\inst{1,2}\orcidlink{0000-0001-9844-8185} \and
Vasileios Argyriou\inst{3}\orcidlink{0000-0003-4679-8049} \and
Ioannis D. Moscholios\inst{4}\orcidlink{0000-0003-3656-277X} \and
Panagiotis Sarigiannidis\inst{1,2}\orcidlink{0000-0001-6042-0355}
}
\authorrunning{G. Tsoumplekas et al.}

\institute{MetaMind Innovations P.C., Kozani, Greece \email{\{gtsoumplekas,isiniosoglou,psarigiannidis\}@metamind.gr} \and Department of Electrical and Computer Engineering, University of Western Macedonia, Kozani, Greece \email{\{isiniosoglou, psarigiannidis\}@uowm.gr} \and Department of Networks and
Digital Media, Kingston University, Kingston upon Thames, UK \email{vasileios.argyriou@kingston.ac.uk} \and Department of Informatics and Telecommunications, University of Peloponnese, Tripolis, Greece \email{idm@uop.gr}}

\maketitle

\begin{abstract}
The increased availability of medical data has significantly impacted healthcare by enabling the application of machine / deep learning approaches in various instances. However, medical datasets are usually small and scattered across multiple providers, suffer from high class-imbalance, and are subject to stringent data privacy constraints. In this paper, the application of a data regularization algorithm, suitable for learning under high class-imbalance, in a federated learning setting is proposed. Specifically, the goal of the proposed method is to enhance model performance for cardiovascular disease prediction by tackling the class-imbalance that typically characterizes datasets used for this purpose, as well as by leveraging patient data available in different nodes of a federated ecosystem without compromising their privacy and enabling more resource sensitive allocation. The method is evaluated across four datasets for cardiovascular disease prediction, which are scattered across different clients, achieving improved performance. Meanwhile, its robustness under various hyperparameter settings, as well as its ability to adapt to different resource allocation scenarios, is verified.

\end{abstract}

\section{Introduction}
\label{Introduction}
The modern era of Big Data has brought about tremendous changes in various domains that impact people's everyday lives, including healthcare. Specifically, the advent of medical software and devices has enabled the generation of large amounts of assorted data regarding patient records. Consequently, such large quantities of medical data have greatly facilitated the application of machine learning (ML) and deep learning (DL) solutions in the medical field. In many cases, these applications aim to complement conventional medical diagnosis methods by reducing the time needed to process large quantities of data, assisting in better decision-making, and enabling the timely prediction of diseases.

However, real-world medical datasets pose several challenges due to their inherent nature~\cite{mazurowski2008training}. One common problem that typically arises in medical datasets for disease classification is that they tend to be highly skewed since most available data refer to healthy individuals while there is only a limited number of data related to patients suffering from a particular disease. This extreme class-imbalance, however, can hinder the practical application of ML methods since these algorithms fail to identify patients with a disease accurately due to the lack of available data for that class.

At the same time, medical datasets are typically small and scattered across different healthcare providers, making it challenging to train ML models, especially when they are highly imbalanced. Combining datasets to train larger models is also challenging due to privacy constraints imposed on medical data. In recent years, federated learning (FL) has gained popularity for such applications due to its inherent characteristics that ensure data privacy by design since it obviates the need for exchanging sensitive data. However, in such critical cases, it is also crucial to consider the computational cost of these decentralized applications and their ability to adapt to different resource allocation scenarios.

This paper aims to tackle data imbalance and privacy considerations in medical datasets by combining imbalanced learning via data regularization with FL. The proposed method is evaluated using four real-world datasets, leading to improved results across all cases compared to methods that do not incorporate the imbalanced and FL criteria. At the same time, it demonstrates robustness under various hyperparameter settings, while it can also maintain high performance under limited available communication resources. The overall contributions of this paper can be summarized as follows:

\begin{itemize}
    \item Proposes the utilization of Balanced-MixUp~\cite{galdran2021balanced} to deal with imbalanced learning for cardiovascular disease prediction using tabular data.
    \item Applies Balanced-MixUp in a FL setting, enhancing model performance across all clients' datasets while also ensuring data privacy of sensitive patient information.
    \item Presents a thorough evaluation of the proposed method under various experimental settings, demonstrating its robustness and resource efficiency.

\end{itemize}

The rest of this paper is organized as follows: the related work is discussed in Section~\ref{relevant_work}, followed by an overview of the methodology in Section~\ref{Methodology}. Section~\ref{Evaluation} provides a comprehensive analysis of the available data and the models' performance under imbalanced learning in a federated setting, and finally, Section~\ref{Conclusions} concludes the paper.

\section{Related Work}
\label{relevant_work}
\subsection{Federated Learning}
FL has increasingly be seen in the medical sector due to its innate attributes, namely privacy-by-design and resource allocation on the edge, but also due to its ability to optimise models by combining the distributed knowledge of decentralised data. This is also the case with the model fusion algorithms employed to merge the decentrally trained models into an optimised global model, that contains the share knowledge from the remote data. In particular, in~\cite{Yaqoob2023} the authors develop an FL system to address the issue of data privacy for Health Service Provider (HSP) systems. The paper utilizes a modified version of the FedMA~\cite{Hongyi2020} model fusion algorithm to ensure the privacy of heart disease data while also showing optimised results for the utilised data. In~\cite{Yifei2023}, the authors propose an improved FL framework that enhances the security of multi-party collaboration in the sensitive context of healthcare data. The work employs this framework to optimise a decentralised AI model for the prediction of Diabetes Mellitus risk, demonstrating enhanced model accuracy, while providing scalability and efficiency. The authors in~\cite{Yaqoob2022} propose a privacy-aware FL framework for heart disease prediction also leveraging the FedMA algorithm. The work employs a proposed Modified Artificial Bee Colony (M-ABC) optimizer at the client end for optimal feature selection of heart disease data, presenting enhanced accuracy of results. This work is further advanced in~\cite{Khan2023} where the authors propose an asynchronous FL (Async-FL) technique for predicting heart diseases. In particular, they employ a temporally weighted aggregation method on the server to enhance the convergence of the global model, using locally trained models from the decentralised nodes. The work shows high model accuracy while addressing privacy concerns and computational efficiency.

\subsection{Imbalanced Learning}

One of the main challenges that arise in classification tasks within the medical domain is the high class-imbalance that typically discerns the available data~\cite{mazurowski2008training}. Over the years, various approaches have been proposed to handle class-imbalance, ranging from simple techniques such as random oversampling / undersampling to more sophisticated ones such as synthetic sampling with data generation, e.g., SMOTE and its variants. In addition to data manipulation techniques, other approaches have focused on the algorithms used to train the classifier on the imbalanced dataset. Some standard methods include cost-sensitive learning~\cite{khan2017cost} and using different weights for the contribution of each sample during training~\cite{zhou2005training}. More recently, various techniques extending data regularization approaches, such as MixUp~\cite{zhang2017mixup}, have also been successfully applied to deal with class-imbalance. Examples of these techniques include Remix~\cite{chou2020remix}, which assigns the synthetic sample label in favor of the minority class, and Balanced-MixUp~\cite{galdran2021balanced}, which introduces a sampling mechanism favoring the creation of synthetic samples near the minority class samples.

\subsection{Cardiovascular Disease Prediction}

While traditionally, cardiovascular disease prediction has been considered a medical task typically performed by physicians, the availability of consolidated data from various patients has enabled the application of ML methods for this task~\cite{miao2020using}. For instance, in~\cite{rahim2021integrated}, an ensemble of K-Nearest Neighbors (KNN) and logistic regression models is used to perform early diagnosis of cardiovascular diseases in the Framingham~\cite{ashish_bhardwaj_2022} and Cleveland~\cite{misc_heart_disease_45} datasets. Ensembling has also been utilized in~\cite{rustamov2023cardiovascular}, where the proposed stacking ensemble model outperforms all individual models in the Framingham dataset, and in~\cite{mienye2020improved}, where an accuracy-based weighted aging classifier ensemble that contains decision trees trained on random splits of the Cleveland and Framingham datasets is proposed. Finally, in~\cite{el2015feature}, integrating different datasets by extracting a shared set of features using decision trees is proposed. The final predictions are then obtained using a decision tree in the total unified dataset. Our approach also focuses on merging different datasets. However, this is done in a FL setting that ensures data privacy of sensitive patient information while, at the same time, we tackle class-imbalance via data regularization. It is also worth noting that while most of the approaches above aim to tackle class-imbalance using SMOTE or its variants, a direct comparison of our approach to these methods would be unjust since the reported metrics in these works refer to the balanced versions of these datasets.

\section{Methodology}
\label{Methodology}
\subsection{Imbalanced Learning with Data Regularization}

To deal with the high class-imbalance that characterizes many cardiovascular disease prediction datasets, Balanced-MixUp~\cite{galdran2021balanced}, a data regularization technique suitable for imbalanced learning, is utilized.

Before delving into the specifics of Balanced-MixUp, it is crucial to provide a brief overview of MixUp~\cite{zhang2017mixup}, which constitutes the basis of the examined method. MixUp is a data regularization technique initially proposed to enhance deep learning performance by reducing overfitting. It relies on composing novel synthetic samples by taking convex combinations of existing training samples and their corresponding labels. In particular, given a dataset $\mathcal{D} = \{(x_i, y_i)\}_{i=1}^N$ with $N$ samples, where $x_i \in \mathbb{R}^M$ is the $i$-th input sample and $y_i$ is its corresponding label, MixUp replaces $\mathcal{D}$ with a novel synthetic dataset $\Tilde{\mathcal{D}} = \{ (\Tilde{x}_k, \Tilde{y}_k) \}_{k=1}^N$, where:

\begin{equation} \label{mixup_eq}
    \Tilde{x}_k = \lambda x_i + (1-\lambda)x_j,\quad \Tilde{y}_k = \lambda y_i + (1-\lambda) y_j
\end{equation}

\noindent In this case, $(x_i, y_i), (x_j, y_j)$ are randomly drawn from $\mathcal{D}$, with $i \neq j$, and $\lambda \in [0,1]$ is drawn from a Beta distribution, $\lambda \sim Beta(\alpha, \alpha)$, with $\alpha>0$.

While MixUp has successfully been applied to various domains, increasing model generalization, one noticeable drawback is that randomly selecting samples to create $\Tilde{\mathcal{D}}$ does not account for any differences in the number of samples in the dataset's classes. Subsequently, Balanced-MixUp was proposed as an extension of MixUp that considers any class-imbalances in $\mathcal{D}$ and induces oversampling of the minority classes within the MixUp formulation. In particular, instead of combining randomly drawn samples from $\mathcal{D}$, Balanced-MixUp combines randomly sampled $(x_i, y_i)$ with $(x_j, y_j)$ that are uniformly sampled from each class. More formally, given $C$ non-overlapping classes in $\mathcal{D}$, each containing $n_c$ samples so that $\sum_{c=0}^{C-1} n_c = N$, the probability of $(x_i, y_i)$ belonging to class $c$ is $p_{c,i} = \frac{n_c}{N}$, while the probability of $(x_j, y_j)$ belonging to class $c$ is $p_{c,j} = \frac{1}{C}$. As a result, for $(x_j, y_j)$, the probability of sampling from a minority class is increased while the probability of sampling from a majority class is reduced compared to $(x_i, y_i)$. This difference in the sampling mechanisms applied to select the original samples in (\ref{mixup_eq}) leads to the creation of novel synthetic samples closer to the areas where the minority class samples lie, consequently increasing model performance in these areas. Additionally, while Balanced-MixUp's $\lambda$ is also drawn from a Beta distribution, the distribution is now parameterized as $\lambda \sim Beta(\alpha,1)$, with $\alpha>0$.

Although Balanced-MixUp was initially proposed for medical image classification, in Section~\ref{Evaluation}, we show that it can also be effectively applied in the context of cardiovascular disease prediction using tabular data.

\subsection{Federated Learning}

\begin{figure}[!th]
    \centering
    \includegraphics[width=0.99\columnwidth]{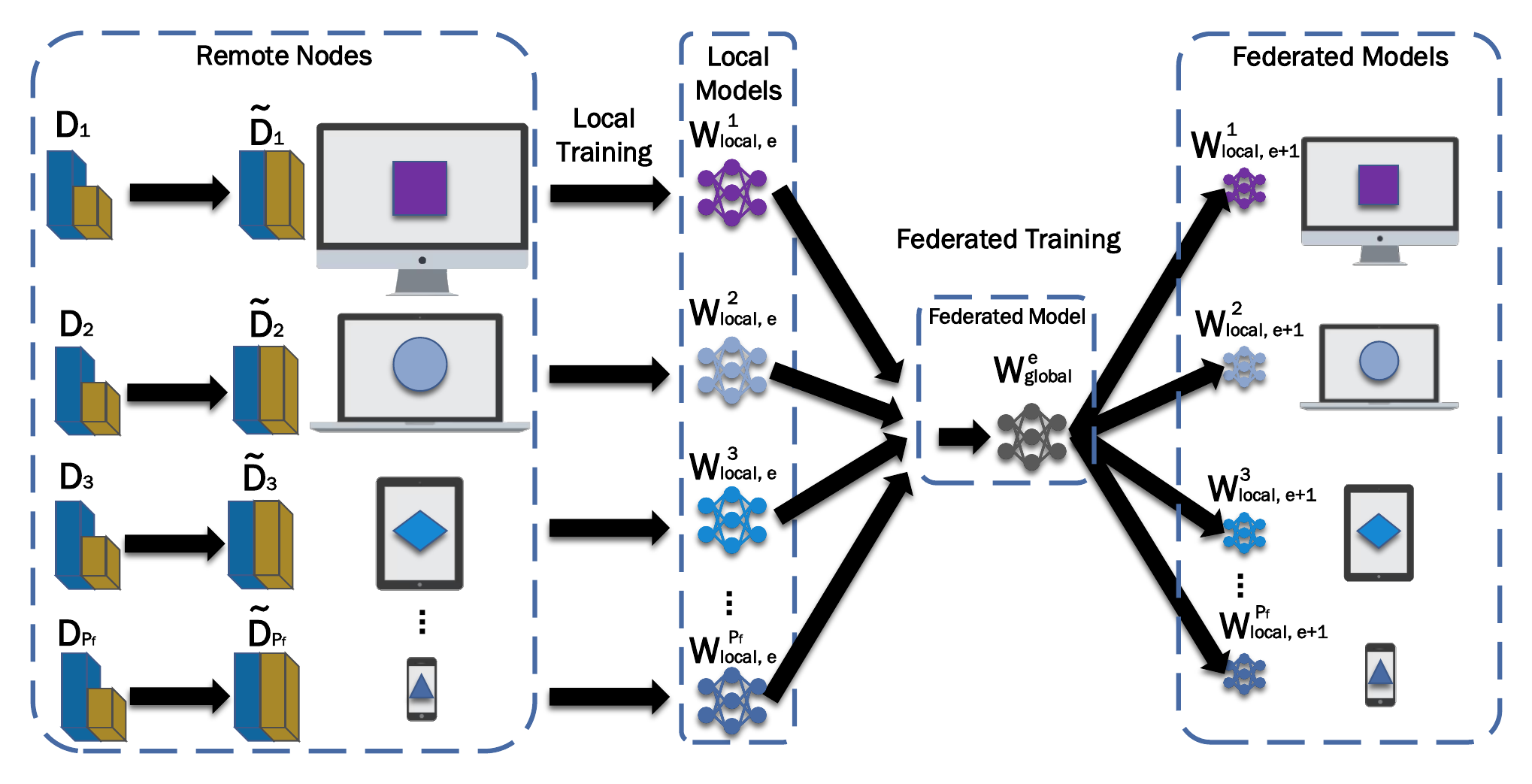}
    \caption{Model training process including employing Balanced-Mixup to deal with class-imbalance and training in a federated learning setting.}
    \label{fig:fl_pipeline}
\end{figure}

Federated Learning (FL) has recently flourished as a widespread tool for distributed machine/deep learning implementation, as it is a technique aimed at privacy preservation. This methodology takes over the orchestration, distribution, learning, and aggregation of deep learning models coming from a large amount of distributed edge devices or remote workers~\cite{Bonawitz2019} that possess local data not available to other devices or the network, as can be seen in Fig.~\ref{fig:fl_pipeline}. Models are trained locally on each device's collected data while the trained model weights are transmitted to a centralized unit where they are subsequently aggregated to produce a mutual global model using a fusion algorithm like Federated Averaging~\cite{McMahan2017}. The fused mutual model is then disbursed back to the edge nodes, updating the previous one. 

Delving deeper into the FL process, the central server disseminates an initial global model \(w_{global}^{0}\) along with metadata about the training procedure to a federated population of \(P_{f} \geq 1\) nodes. Each node holds a set of local data \(\mathcal{D}_p\), with \(p=1,2,..,P_f\), and a local model \(w_{local,0}^{p}\). After transforming each node's dataset to $\Tilde{\mathcal{D}}_p$ using Balanced-MixUp, the distributed models are then trained on the transformed datasets, and the model weights \(w_{global}^{e}\) are retrieved by the central server to be fused utilizing the Federated Averaging algorithm. Here $e=1,2,...,E_{FL}$ refers to the current communication round and $E_{FL}$ is the total number of communication rounds. After each communication round, a new global model \(w_{global}^{e}\)~\cite{Siniosoglou2022} containing the newly collected knowledge is produced based on:

\begin{equation}
w_{global}^{e} = \frac{1}{\sum _{p=1}^{P_f} \Tilde{\mathcal{D}}_{p}} \sum_{p=1}^{P_f} \Tilde{\mathcal{D}}_{p}w_{local,e}^{p}
\label{FederatedAveraging}
\end{equation}

\noindent where, $w_{global}^{e}$ is the global model on the \(e_{th}\) communication round and \(w_{local,e}^{p}\) is the $p$-th remote model at that round.

\section{Experimental Results}
\label{Evaluation}
\subsection{Experimental Setting} \label{exp_setting}

Regarding imbalanced learning evaluation, in each of the following experiments, apart from Balanced-MixUp, denoted as \textit{Bal-MixUp}, standard MixUp, denoted as \textit{MixUp}, and training without any data regularization, denoted as \textit{No MixUp}, are also examined for comparison purposes. Additionally, the advantages of leveraging a FL approach, denoted as \textit{FL} in subsequent experiments, are demonstrated by comparing the relevant results with those obtained in a local training setting, denoted as \textit{Local}. To ensure a fair comparison among these methods, we use the same multilayer perceptron (MLP) network model consisting of two hidden layers with a size 128. The models are optimized using Adam~\cite{kingma2014adam}, and the batch size is set to 24. For \textit{No MixUp} and \textit{MixUp}, we apply a learning rate of 0.004, while for \textit{Bal-MixUp} the learning rate is set to 0.034. For \textit{MixUp}, optimal results are obtained for $\alpha$ = 0.1, while for \textit{Bal-MixUp} optimal performance is achieved with $\alpha$ = 0.3. Finally, in the local training scenario, each model is trained for 100 epochs. In the FL scenario, we set the number of communication rounds to 5. However, to ensure a fair comparison between the two scenarios, the number of local epochs in the FL scenario is reduced to 20 so that the number of minibatch gradient calculations remains the same across both settings. Due to hardware constraints, each reported result corresponds to a single experiment run.

\subsection{Datasets} \label{datasets}

For the following experiments, four different tabular datasets related to cardiovascular disease prediction are examined. Specifically, these data represent different patients' records, and the problem is formulated as a binary classification task where the goal is to predict whether a given patient will suffer from cardiovascular disease after ten years. For all datasets, an 80/20 train-test split ratio was used. The specific datasets are:

\begin{itemize}
    \item Framingham~\cite{ashish_bhardwaj_2022}: Consists of 4240 health record samples from different patients, of which $15.19 \%$ have suffered from cardiovascular disease after ten years.

    \item Cleveland~\cite{misc_heart_disease_45}: Consists of 282 records from different patients, with $44.33 \%$ suffering from cardiovascular disease after ten years.

    \item Long Beach~\cite{misc_heart_disease_45}: It contains 200 samples, and $74.5 \%$ of them are classified as suffering from cardiovascular disease after ten years.

    \item Switzerland~\cite{misc_heart_disease_45}: It consists of 123 different patient samples, of which $95.5 \%$ have been diagnosed with cardiovascular disease after ten years.
\end{itemize}

\begin{figure}
 \begin{center}
  \includegraphics[width=0.72\columnwidth]{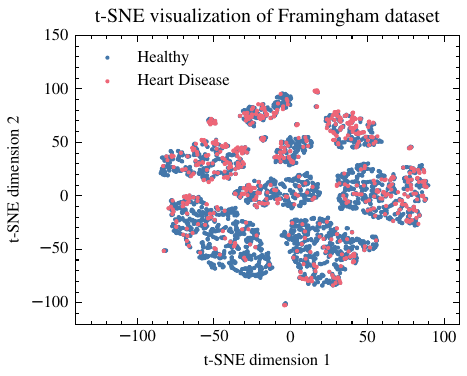}
 \end{center}
 \caption{Two-dimensional t-SNE representations of Framingham's samples.}
 \label{framingham_tsne}
\end{figure}

\noindent Overall, it is evident that all examined datasets are small and demonstrate high class-imbalance, rendering model training in them a nontrivial task. The difficulty of training efficient models in these datasets is also illustrated in Fig.~\ref{framingham_tsne}, which shows that there is no clear decision boundary for the t-SNE representations of Framingham's samples. Additionally, while all four of the aforementioned datasets contain tabular data, including demographic and behavioral patient characteristics as well as patients' medical history and current medical conditions, their included features are not identical. Consequently, to allow their utilization in a homogeneous FL setting, we only consider a subset of 10 features available in all four datasets. Table~\ref{features} contains a short overview of these features. Finally, before being used for training and evaluation, all features are normalized, and mean value interpolation is used to fill any missing values.

\begin{table}[]
\centering
\caption{Description of the shared features across the examined datasets.}
\label{features}
\begin{tabular}{p{2cm}p{3cm}l}
    \toprule
    Feature & Type & Description \\
    \midrule
    age & Continuous & Patient age \\
    male & Binary & Patient gender  \\
    hyp & Binary & Patient suffers from hypertension \\
    smoker & Binary & Patient is a smoker \\
    cigsperday & Continuous & Number of cigarettes smoked per day \\
    diabetes & Binary & Patient has diabetes \\
    chol & Continuous & Total cholesterol level \\
    heartRate & Continuous & Patient heart rate \\
    sysBP & Continuous & Systolic blood pressure \\
    diaBP & Continuous & Diastolic blood pressure \\
    \bottomrule
\end{tabular} 
\end{table}

\subsection{Evaluation Metrics} \label{metrics}

Since the problem at hand is formulated as a binary classification task, the following metrics have been leveraged to evaluate model performance:

\begin{itemize}
    \item {\textbf{Binary Cross-Entropy:} It is used as the loss that is minimized during training, but we also report its values during model evaluation. It is formulated as:

    \begin{equation}
        \mathcal{L}_{BCE} = -\frac{1}{N} \sum_{i=1}^N (y_i \log(\hat{y_i}) + (1-y_i)\log(1-\hat{y}_i))
    \end{equation}

    where $y_i$ are the true and $\hat{y}_i$ are the predicted values for $N$ test data points.}

    \item {\textbf{Accuracy:} Accuracy refers to the fraction of correct model predictions over all predictions made. Given a confusion matrix containing the number of True Positive (TP), True Negative (TN), False Positive (FP), and False Negative (FN) values, accuracy (\textit{Acc}) can be formulated as:

    \begin{equation}
        Acc = \frac{TP+TN}{TP+TN+FP+FN}
    \end{equation}
    }

    \item {\textbf{F-Score:} It is defined as the harmonic mean between the model's precision and recall, consolidating and striking a balance between these two metrics. It can also be formulated as:

    \begin{equation}
        \textit{F-Score} = \frac{2TP}{2TP+FP+FN}
    \end{equation}
    }
\end{itemize}

It is worth noting that due to the high class-imbalance that discerns the examined datasets, \textit{Acc.} can lead to misleading results. Consequently, our primary focus is towards models that achieve higher \textit{F-Score} values, which is inherently less biased under class-imbalance.

\begin{table}[h]
\centering
\caption{Model performance metrics for each dataset in the local training and federated learning setting.}
\label{main_results_table}
\resizebox{0.99\columnwidth}{!}{%
\begin{tabular}{p{2.5cm}p{3,5cm}cccccc}
\toprule
\multicolumn{1}{l}{\multirow{2}{*}{\textbf{Dataset}}} & \multicolumn{1}{l}{\multirow{2}{*}{\textbf{Model}}} & \multicolumn{3}{c}{\textbf{Local}} & \multicolumn{3}{c}{\textbf{FL, 5 rounds}} \\
\cmidrule(r){3-5}\cmidrule(l){6-8}
    & & \textbf{Loss} & \textbf{Acc.} & \textbf{F-Score} & \textbf{Loss} & \textbf{Acc.} & \textbf{F-Score} \\
    \midrule
    \multicolumn{1}{l}{\multirow{3}{*}{\textbf{Framingham}}}
        & No MixUp & 0.408 & 84.52 & 50.63 & \textbf{0.396} & \textbf{84.64} & 45.84  \\
        & MixUp ($\alpha=0.1$) & \textbf{0.402} & \textbf{84.76} & 47.38 & 0.398 & \textbf{84.64} & 49.41  \\
        & Bal-MixUp ($\alpha=0.3$) & 0.585 & 69.52 & \textbf{56.95} & 0.537 & 74.05 & \textbf{59.08} \\
    \midrule
    \multicolumn{1}{l}{\multirow{3}{*}{\textbf{Cleveland}}}
        & No MixUp & 0.720 & 62.50 & 61.90 & 0.670 & 54.17 & 35.14  \\
        & MixUp ($\alpha=0.1$) & 0.581 & \textbf{72.92} & \textbf{72.62} & 0.632 & 58.33 & 49.58  \\
        & Bal-MixUp ($\alpha=0.3$) & \textbf{0.560} & 66.67 & 66.61 & \textbf{0.582} & \textbf{70.83} & \textbf{70.78} \\
    \midrule
    \multicolumn{1}{l}{\multirow{3}{*}{\textbf{Long Beach}}}
        & No MixUp & 0.685 & 62.50 & 46.93 & 0.845 & 20.83 & 17.24  \\
        & MixUp ($\alpha=0.1$) & \textbf{0.623} & 66.67 & 49.47 & 0.720 & 50.00 & 48.57  \\
        & Bal-MixUp ($\alpha=0.3$) & 0.730 & \textbf{75.00} & \textbf{69.75} & \textbf{0.555} & \textbf{79.17} & \textbf{70.52} \\
    \midrule
    \multicolumn{1}{l}{\multirow{3}{*}{\textbf{Switzerland}}}
        & No MixUp & 0.696 & \textbf{83.33} & \textbf{45.45} & 0.851 & 12.50 & 11.11  \\
        & MixUp ($\alpha=0.1$) & \textbf{0.635} & \textbf{83.33} & \textbf{45.45} & 0.729 & 45.83 & 40.80  \\
        & Bal-MixUp ($\alpha=0.3$) & 0.665 & 70.83 & 41.46 & \textbf{0.440} & \textbf{91.67} & \textbf{72.73} \\
    \bottomrule
\end{tabular} 
}
\end{table}

\subsection{Experimental Results} \label{exp_results}

\subsubsection{Main Results. } Table~\ref{main_results_table} shows the performance results of the three examined models for both local and FL settings in each dataset. Initially, it is worth noting that the loss and \textit{Acc.} values fail to provide a clear view of which model performs best, especially in the local setting where, depending on the dataset, optimal results are either obtained using \textit{MixUp} or \textit{Bal-MixUp}. However, using a data regularization technique in all four datasets leads to better or at least equivalent results compared to \textit{No-MixUp}. This is also true for \textit{F-Score} in the local training setting. In the federated setting, it is evident that \textit{Bal-MixUp} outperforms all other methods, demonstrating the best \textit{F-Score} in all four datasets. Although \textit{Bal-MixUp}'s loss and \textit{Acc.} are worse than those of the other two models for the Framingham dataset, these values can be misleading due to the high class-imbalance. As a result, \textit{F-Score} is used as a more reliable performance indicator. Additionally, when moving to the FL setting, while \textit{No MixUp} and \textit{MixUp} show worse \textit{F-Score} performance, \textit{Bal-MixUp}'s performance is improved, showing a $2.13\%$ improvement for Framingham, $4.17\%$ for Cleveland, $0.77\%$ for Long Beach, and $31.27\%$ for Switzerland. Overall, it is clear that the combination of using data regularization techniques, especially when dealing with high class-imbalance, as in \textit{Bal-MixUp}, as well as leveraging additional data during training, as in the case of FL, can improve overall performance in all datasets involved.

\begin{figure}[h]
        \centering
        \begin{subfigure}[b]{0.475\textwidth}
            \centering
            \includegraphics[width=\textwidth]{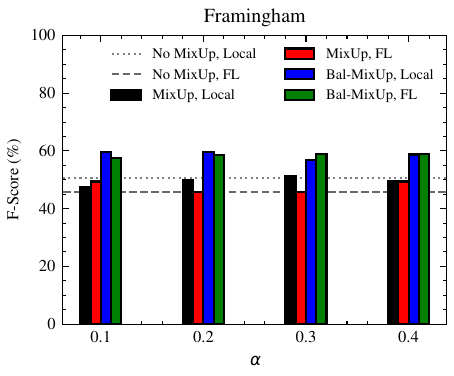}
            \caption[]
            {{\small Framingham Dataset}}    
            \label{framingham_alpha}
        \end{subfigure}
        \hfill
        \begin{subfigure}[b]{0.475\textwidth}  
            \centering 
            \includegraphics[width=\textwidth]{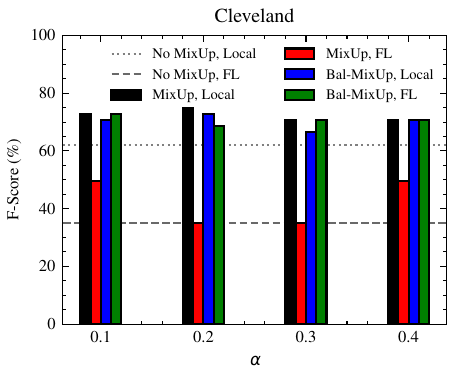}
            \caption[]
            {{\small Cleveland Dataset}}    
            \label{cleveland_alpha}
        \end{subfigure}
        \vskip\baselineskip
        \begin{subfigure}[b]{0.475\textwidth}   
            \centering 
            \includegraphics[width=\textwidth]{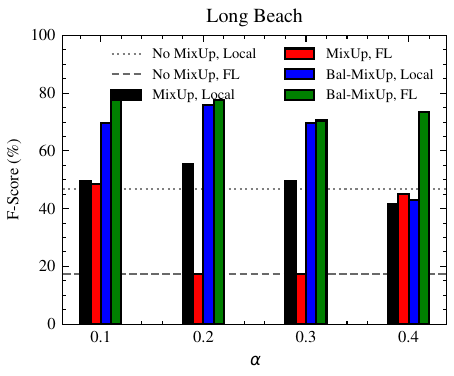}
            \caption[]
            {{\small Long Beach Dataset}}    
            \label{long_bach_alpha}
        \end{subfigure}
        \hfill
        \begin{subfigure}[b]{0.475\textwidth}   
            \centering 
            \includegraphics[width=\textwidth]{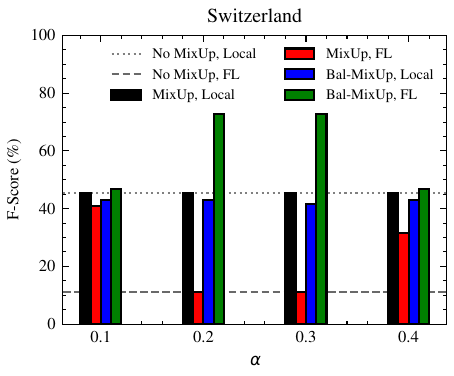}
            \caption[]
            {{\small Switzerland Dataset}}    
            \label{switzerland_alpha}
        \end{subfigure}
        \caption[]
        {\small F-Score of examined methods for varying values of $\alpha$ in each dataset.} 
        \label{varying_alpha}
    \end{figure}

\subsubsection{Effect of data regularization. } In general, the $\alpha$ hyperparameter used in \textit{MixUp} and \textit{Bal-MixUp} influences the shape of the Beta distribution from which the $\lambda$ coefficients in (\ref{mixup_eq}) are obtained, with lower $\alpha$ values leading to synthetic samples that are more likely to be close to the original samples used. Following~\cite{zhang2017mixup, galdran2021balanced} we test different $\alpha$ values in the range of $[0.1, 0.4]$. Fig.~\ref{varying_alpha} illustrates the effect of $\alpha$ on model performance in the \textit{local} and \textit{FL} setting. In the Framingham and Long Beach datasets, \textit{Bal-MixUp} consistently outperforms all other methods. As for the Switzerland dataset, while \textit{Bal-MixUp, FL} still outperforms all other methods, \textit{Bal-MixUp, Local}'s performance is not optimal, pinpointing the need for leveraging additional data in an FL setting, especially when data is scarce. Finally, in the Cleveland dataset, the best results are achieved using \textit{MixUp, Local}. However, its performance on the rest of the datasets is generally worse compared to \textit{Bal-MixUp}. On the other hand, \textit{Bal-MixUp}'s performance in the Cleveland dataset is still on par compared to \textit{MixUp}. Consequently, \textit{Bal-MixUp} demonstrates the best overall performance irrespective of the $\alpha$ value chosen.

\begin{figure}[h]
        \centering
        \begin{subfigure}[b]{0.475\textwidth}
            \centering
            \includegraphics[width=\textwidth]{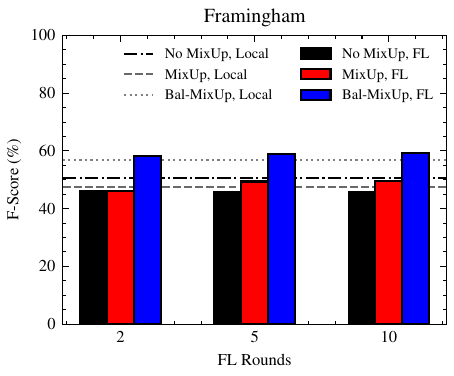}
            \caption[]
            {{\small Framingham Dataset}}    
            \label{framingham_fl}
        \end{subfigure}
        \hfill
        \begin{subfigure}[b]{0.475\textwidth}  
            \centering 
            \includegraphics[width=\textwidth]{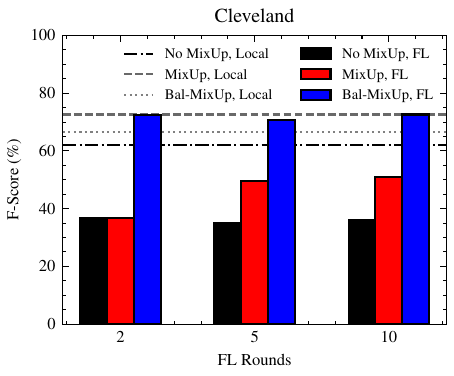}
            \caption[]
            {{\small Cleveland Dataset}}    
            \label{cleveland_fl}
        \end{subfigure}
        \vskip\baselineskip
        \begin{subfigure}[b]{0.475\textwidth}   
            \centering 
            \includegraphics[width=\textwidth]{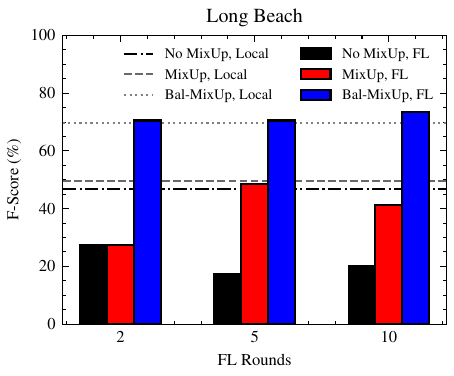}
            \caption[]
            {{\small Long Beach Dataset}}    
            \label{long_beach_fl}
        \end{subfigure}
        \hfill
        \begin{subfigure}[b]{0.475\textwidth}   
            \centering 
            \includegraphics[width=\textwidth]{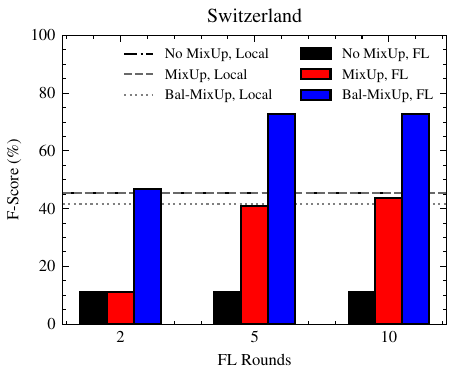}
            \caption[]
            {{\small Switzerland Dataset}}    
            \label{switzerland_dataset}
        \end{subfigure}
        \caption[]
        {\small F-Score of examined methods for a varying number of communication rounds in each dataset.} 
        \label{varying_fl_rounds}
    \end{figure}

\subsubsection{Effect of federated learning. } While increasing the number of communication rounds in the FL setting can lead to better generalization performance across all clients, it can also lead to increased communication overhead, negatively impacting the model's time efficiency. In this set of experiments, we examine how varying the number of communication rounds in FL affects performance. To ensure a fair comparison across different numbers of communication rounds, we keep the number of gradient calculations constant across all settings by setting $E_{FL} \times E_{local} = 100$, where $E_{FL}$ is the number of communication rounds and $E_{local}$ is the number of local epochs within each round. Fig.~\ref{varying_fl_rounds} illustrates the \textit{F-Score} achieved by each method for 2, 5, and 10 communication rounds. Initially, \textit{Bal-MixUp} outperforms all other methods in all datasets except Cleveland. However, its performance is still on par with the best-performing model, \textit{MixUp, Local}. Additionally, \textit{Bal-MixUp} exhibits robustness in its performance even when the number of communication rounds is reduced, which is desirable as it allows reducing the communication rounds without negatively impacting the model's performance. However, this is not true for the rest of the models, which are more sensitive to this hyperparameter. Specifically, \textit{MixUp}'s performance generally decreases when reducing the number of communication rounds, while it is augmented when increasing the number of communication rounds. However, it still underperforms compared to \textit{Bal-MixUp}. Overall, the positive impact of the synergies between imbalanced learning via data regularization and FL is also manifest in this setting by the superior performance of the \textit{Bal-MixUp} model.

\section{Conclusions}
\label{Conclusions}
One of the most challenging points when deploying ML models in healthcare domain applications is dealing with the imbalance that typically characterizes medical data. At the same time, the need to ensure data privacy of sensitive patient information has led to the increased utilization of FL approaches in this domain. One medical application where meeting both these requirements is crucial is cardiovascular disease prediction. In this paper, we propose employing Balanced-MixUp, a data-regularization method suitable for dealing with class imbalance, in a FL setting that enforces data privacy by design for cardiovascular disease prediction. The proposed method was evaluated in a realistic setting with multiple nodes, outperforming all methods that failed to address both imbalanced and FL needs. Finally, further experiments were conducted to demonstrate the proposed method's robustness under different hyperparameter values and resource efficiency under limited resource allocation in the federated setting.

\begin{credits}
\subsubsection{\ackname} This project has received funding from the European Union’s Horizon Europe research and innovation programme under grant agreement No. \\101095435 (REALM).

\subsubsection{\discintname}
The authors have no competing interests to declare that are relevant to the content of this article.
\end{credits}

\bibliographystyle{splncs04}
\bibliography{bibliography}

\begin{thebibliography}{10}
\providecommand{\url}[1]{\texttt{#1}}
\providecommand{\urlprefix}{URL }
\providecommand{\doi}[1]{https://doi.org/#1}

\bibitem{ashish_bhardwaj_2022}
Bhardwaj, A.: Framingham heart study dataset (2022). \doi{10.34740/KAGGLE/DSV/3493583}

\bibitem{Bonawitz2019}
Bonawitz, K., Eichner, H., Grieskamp, W., Huba, D., Ingerman, A., Ivanov, V., Kiddon, C., Konečný, J., Mazzocchi, S., McMahan, H., Overveldt, T., Petrou, D., Ramage, D., Roselander, J.: Towards federated learning at scale: System design  (02 2019)

\bibitem{chou2020remix}
Chou, H.P., Chang, S.C., Pan, J.Y., Wei, W., Juan, D.C.: Remix: rebalanced mixup. In: Computer Vision--ECCV 2020 Workshops: Glasgow, UK, August 23--28, 2020, Proceedings, Part VI 16. pp. 95--110. Springer (2020)

\bibitem{el2015feature}
El-Bialy, R., Salamay, M.A., Karam, O.H., Khalifa, M.E.: Feature analysis of coronary artery heart disease data sets. Procedia Computer Science  \textbf{65},  459--468 (2015)

\bibitem{galdran2021balanced}
Galdran, A., Carneiro, G., Gonz{\'a}lez~Ballester, M.A.: Balanced-mixup for highly imbalanced medical image classification. In: Medical Image Computing and Computer Assisted Intervention--MICCAI 2021: 24th International Conference, Strasbourg, France, September 27--October 1, 2021, Proceedings, Part V 24. pp. 323--333. Springer (2021)

\bibitem{misc_heart_disease_45}
Janosi, A., Steinbrunn, W., Pfisterer, M., Detrano, R.: {Heart Disease}. UCI Machine Learning Repository (1988), {DOI}: https://doi.org/10.24432/C52P4X

\bibitem{Khan2023}
Khan, M.A., Alsulami, M., Yaqoob, M.M., Alsadie, D., Saudagar, A.K.J., AlKhathami, M., Farooq~Khattak, U.: Asynchronous federated learning for improved cardiovascular disease prediction using artificial intelligence. Diagnostics  \textbf{13}(14) (2023). \doi{10.3390/diagnostics13142340}

\bibitem{khan2017cost}
Khan, S.H., Hayat, M., Bennamoun, M., Sohel, F.A., Togneri, R.: Cost-sensitive learning of deep feature representations from imbalanced data. IEEE transactions on neural networks and learning systems  \textbf{29}(8),  3573--3587 (2017)

\bibitem{kingma2014adam}
Kingma, D.P., Ba, J.: Adam: A method for stochastic optimization. arXiv preprint arXiv:1412.6980  (2014)

\bibitem{mazurowski2008training}
Mazurowski, M.A., Habas, P.A., Zurada, J.M., Lo, J.Y., Baker, J.A., Tourassi, G.D.: Training neural network classifiers for medical decision making: The effects of imbalanced datasets on classification performance. Neural networks  \textbf{21}(2-3),  427--436 (2008)

\bibitem{McMahan2017}
McMahan, B., Moore, E., Ramage, D., Hampson, S., y~Arcas, B.A.: Communication-efficient learning of deep networks from decentralized data. In: Artificial intelligence and statistics. pp. 1273--1282. PMLR (2017)

\bibitem{miao2020using}
Miao, L., Guo, X., Abbas, H.T., Qaraqe, K.A., Abbasi, Q.H.: Using machine learning to predict the future development of disease. In: 2020 international conference on UK-China emerging technologies (UCET). pp.~1--4. IEEE (2020)

\bibitem{mienye2020improved}
Mienye, I.D., Sun, Y., Wang, Z.: An improved ensemble learning approach for the prediction of heart disease risk. Informatics in Medicine Unlocked  \textbf{20},  100402 (2020)

\bibitem{rahim2021integrated}
Rahim, A., Rasheed, Y., Azam, F., Anwar, M.W., Rahim, M.A., Muzaffar, A.W.: An integrated machine learning framework for effective prediction of cardiovascular diseases. IEEE Access  \textbf{9},  106575--106588 (2021)

\bibitem{rustamov2023cardiovascular}
Rustamov, Z., Rustamov, J., Sultana, M.S., Ywei, J., Balakrishnan, V., Zaki, N.: Cardiovascular disease prediction using ensemble learning techniques: A stacking approach. In: 2023 19th IEEE International Colloquium on Signal Processing \& Its Applications (CSPA). pp. 93--98. IEEE (2023)

\bibitem{Siniosoglou2022}
Siniosoglou, I., Argyriou, V., Lagkas, T., Moscholios, I., Fragulis, G., Sarigiannidis, P.: Unsupervised bias evaluation of dnns in non-iid federated learning through latent micro-manifolds. In: IEEE INFOCOM 2022 - IEEE Conference on Computer Communications Workshops (INFOCOM WKSHPS). pp.~1--6 (2022). \doi{10.1109/INFOCOMWKSHPS54753.2022.9798157}

\bibitem{Yifei2023}
Su, Y., Huang, C., Zhu, W., Lyu, X., Ji, F.: Multi-party diabetes mellitus risk prediction based on secure federated learning. Biomedical Signal Processing and Control  \textbf{85},  104881 (2023). \doi{https://doi.org/10.1016/j.bspc.2023.104881}

\bibitem{Hongyi2020}
Wang, H., Yurochkin, M., Sun, Y., Papailiopoulos, D., Khazaeni, Y.: Federated learning with matched averaging (2020)

\bibitem{Yaqoob2023}
Yaqoob, M.M., Nazir, M., Khan, M.A., Qureshi, S., Al-Rasheed, A.: Hybrid classifier-based federated learning in health service providers for cardiovascular disease prediction. Applied Sciences  \textbf{13}(3) (2023). \doi{10.3390/app13031911}

\bibitem{Yaqoob2022}
Yaqoob, M.M., Nazir, M., Yousafzai, A., Khan, M.A., Shaikh, A.A., Algarni, A.D., Elmannai, H.: Modified artificial bee colony based feature optimized federated learning for heart disease diagnosis in healthcare. Applied Sciences  \textbf{12}(23) (2022). \doi{10.3390/app122312080}

\bibitem{zhang2017mixup}
Zhang, H., Cisse, M., Dauphin, Y.N., Lopez-Paz, D.: mixup: Beyond empirical risk minimization. arXiv preprint arXiv:1710.09412  (2017)

\bibitem{zhou2005training}
Zhou, Z.H., Liu, X.Y.: Training cost-sensitive neural networks with methods addressing the class imbalance problem. IEEE Transactions on knowledge and data engineering  \textbf{18}(1),  63--77 (2005)

\end{thebibliography}

\end{document}